\documentclass{article}

\usepackage{PRIMEarxiv}

\usepackage[utf8]{inputenc} 
\usepackage[T1]{fontenc}    
\usepackage[hidelinks]{hyperref}       
\usepackage{url}            
\usepackage{booktabs}       
\usepackage{amsfonts}       
\usepackage{nicefrac}       
\usepackage{microtype}      
\usepackage{fancyhdr}       
\usepackage{graphicx}       

\usepackage{mathptmx} 
\usepackage{amsmath} 
\usepackage{amssymb}  
\usepackage{amsthm}
\usepackage{subfig}
\usepackage{subcaption}
\usepackage{soul}
\usepackage{balance}
\usepackage{array}
\usepackage{multirow}
\usepackage{cleveref}
\usepackage[labelfont=bf, justification=justified, format=plain]{caption} 

\title{Submeter-level Land Cover Mapping of Japan
}

\author{Naoto Yokoya\thanks{\textit{Corresponding author:} yokoya@k.u-tokyo.ac.jp} \\
  Graduate School of Frontier Sciences, The University of Tokyo, Chiba, 277-8561, Japan \\
  RIKEN Center for Advanced Intelligence Project (AIP), Tokyo, Japan\\
  \texttt{yokoya@k.u-tokyo.ac.jp} \\
  \And
  {Junshi Xia}, {Clifford Broni-Bediako} \\
 RIKEN Center for Advanced Intelligence Project (AIP), Tokyo, Japan\\
 \texttt{\{junshi.xia, clifford.broni-bediako\}@riken.jp} \\
}

\begin{document}
\maketitle

\begin{abstract}
Deep learning has shown promising performance in submeter-level mapping tasks; however, the annotation cost of submeter-level imagery remains a challenge, especially when applied on a large scale. In this paper, we present the first submeter-level land cover mapping of Japan with eight classes, at a relatively low annotation cost. We introduce a human-in-the-loop deep learning framework leveraging OpenEarthMap, a recently introduced benchmark dataset for global submeter-level land cover mapping, with a U-Net model that achieves national-scale mapping with a small amount of additional labeled data. By adding a small amount of labeled data of areas or regions where a U-Net model trained on OpenEarthMap clearly failed and retraining the model, an overall accuracy of 80\% was achieved, which is a nearly 16 percentage point improvement after retraining. Using aerial imagery provided by the Geospatial Information Authority of Japan, we create land cover classification maps of eight classes for the entire country of Japan. Our framework, with its low annotation cost and high-accuracy mapping results, demonstrates the potential to contribute to the automatic updating of national-scale land cover mapping using submeter-level optical remote sensing data. The mapping results will be made publicly available.
\end{abstract}


\section{Introduction}\label{sec:1}
Land cover mapping is the process of categorizing and mapping the different types of land cover or land use, such as forests, urban areas, water bodies, and agricultural fields, that can be found in a given geographic area. It involves analyzing remote sensing data, such as satellite imagery, and assigning a land cover class label to each pixel. The goal is to provide valuable information about the spatial distribution and composition of land cover types, which is helpful for various applications, including environmental monitoring, urban planning, agriculture monitoring, and disaster management. 

Supported by advancements in data accessibility and data analysis techniques, significant efforts have been made in the creation of global-scale land cover maps. For example, the MODIS Land Cover Type Product (MCD12Q1) is a land cover map product with a resolution of 500 meters and 17 classes, created from the MODIS satellite data using techniques such as unmixing and machine learning~\cite{Friedl2019-oq}. It includes yearly maps from 2001 to 2020 that have been made publicly available. Over the past decade, the development of medium-resolution land cover maps, such as GlobeLand30 \cite{chen2015global} and FROM-GLC30~\cite{gong2013finer}, using Landsat satellite data, has significantly advanced. With the introduction of Sentinel-2 satellites, the resolution of global land cover maps has been increased to 10 meters. Representative 10-meter resolution global land cover products include FROM-GLC10~\cite{chen2019stable}, Esri Land Cover Map, and WorldCover 10m~\cite{zanaga_daniele_2021_5571936}. In data processing, the mainstream approach involves using machine learning to classify the time-series spectral information at each pixel. Machine learning techniques such as maximum likelihood, decision trees, random forests~\cite{breiman2001random}, and support vector machines~\cite{cortes1995support}, among others, have been widely used for global land cover classification using pixel-wise features on meter-level satellite image data.

Each country or region (e.g., Europe) has been developing detailed and accurate land cover maps. For instance, the European Union’s CORINE Land Cover Map utilizes more reliable labels from field surveys, enabling a highly accurate and detailed classification with 44 classes~\cite{bossard2000corine, weber2007implementation}. The European Urban Atlas~\cite{Montero2014} produced by the GMES Urban Services project covers more than 300 major European cities using SPOT 5 satellite imagery. The cities are mapped using 20 classes, of which 17 urban classes have a minimum mapping unit (MMU) of 0.25 ha and 3 non-urban classes of 1 ha. In Japan, Deng~\textit{et al.} produced a land cover map of 6 classes for the Gunma Prefecture by combining Landstat-5 TM, SPOT, and aerial imagery~\cite{Deng2003MappingLC}. The Japan Aerospace Exploration Agency (JAXA) and the University of Tsukuba have regularly maintained land cover maps with a ground sampling distance of 10 to 30 meters~\cite{jaxaDatasetALOSEORC}. There are three versions for the time periods of 2006–2011, 2014–2016, and 2018–2020, using time-series multispectral data from ALOS, Landsat-8, and Sentinel-2, respectively. For classification methods, kernel density estimation was used for the first and second versions; the latest version was made using convolutional neural networks and exploiting spectral-temporal features. Sharma~\textit{et al.} adopted a random forest technique with bootstrap aggregating (bagging) to produce the Japan 30-meter resolution land cover map with 7 land cover types, using Landsat-8 Operational Land Imager and Thermal Infrared Sensor scenes over Japan from 2013 to 2015~\cite{Sharma2016}. 

The improvement in resolution is crucial for obtaining the more detailed spatial information that is essential for environmental conservation and land use planning. The Chesapeake Bay Program in the US created 2013/14 and 2017/18 high-resolution land use and land cover maps that classify 54 detailed classes at 1-meter resolution by rule-based classification using National Agriculture Imagery Program (NAIP) images and above-ground height information derived from LiDAR data~\cite{chesapeakeconservancyChesapeakeProgram2013-14,chesapeakeconservancyChesapeakeProgram2017-18}. Spatial patterns and features are informative for land cover classification of high-resolution imagery, and deep learning is a powerful tool for automating map production~\cite{sirko2021continental}. For example, the 1-meter resolution database named UrbanWatch containing images from 22 major cities across the continental United States, was generated by a teacher-student deep learning method using 2014 to 2017 NAIP airborne imagery~\cite{ZHANG2022113106}. \cite{shiUGS-1m2023} adopted a deep learning method consisting of a generator and a discriminator, known as generative adversarial network (GAN), to generate 1-meter urban green space (UGS-1m) maps of 34 major cities and areas in China with global urban boundaries~\cite{Li2020GUB} and Google Earth imagery~\cite{gorelick2017google}. The GlobalUrbanNet automatic multi-city mapping and analysis framework introduced by \cite{zhong2023GUN} adopted both a convolutional neural network and a vision transformer to generate land cover and land use mapping of global cities automatically, at a 0.5-m resolution based on open-source data from OpenStreetMap and very high-resolution images that were purchased or collected from various open-source sites.

The generalization performance of deep learning models depends on the quantity and quality of labels. Various data has been developed for submeter-level image segmentation over the past decade, mostly for building detection~\cite{van2018spacenet,OpenCitiesAI}. However, datasets for land cover mapping are much less available than those for building detection. Benchmark datasets for land cover mapping at the submeter level include DeepGlobe~\cite{demir2018deepglobe} and LoveDA~\cite{wang2021loveda}, but there is little diversity of the regions and countries covered by these datasets, and the geographic coverage of models trained on them is limited. It has been difficult to obtain high-resolution training data covering the entire globe due to the high cost of image acquisition and labeling. OpenEarthMap solved this problem, improving global generalization performance and eliminating geographic unfairness in model performance~\cite{xia2023openearthmap}. The data was collected from existing benchmark data as well as from open data in data-poor regions in a geographically balanced manner, resulting in a consistent 8-class labeling of imagery covering 44 countries and 97 regions. The labeling was very fine-grained and resulted in the construction of a submeter-level land cover mapping model that is globally applicable.

This paper presents a submeter-level land cover mapping of the entirety of Japan using OpenEarthMap. In particular, we present a human-in-the-loop deep learning framework for building a customized model when there are unseen images that are out of the distribution of the OpenEarthMap dataset. The country-scale submeter-level aerial imagery used in this study is a collection of images from multiple time periods, so it contains many ``wild'' cases, such as image mosaic boundaries, heavy shadows, and low-quality old images. Since there are many cases where simply applying the OpenEarthMap pretrained model does not work, we created a Japan-specific model by additionally labeling the scenes where the estimation of the model is poor, and retraining the model with the newly labeled data. We confirm the high performance of mapping through the use of both numerical and visual evaluations. Our map will be made publicly available.


\section{Materials and Methods}\label{sec:2}
\begin{figure}[t]
	\centering 
	\includegraphics[width=0.8\textwidth]{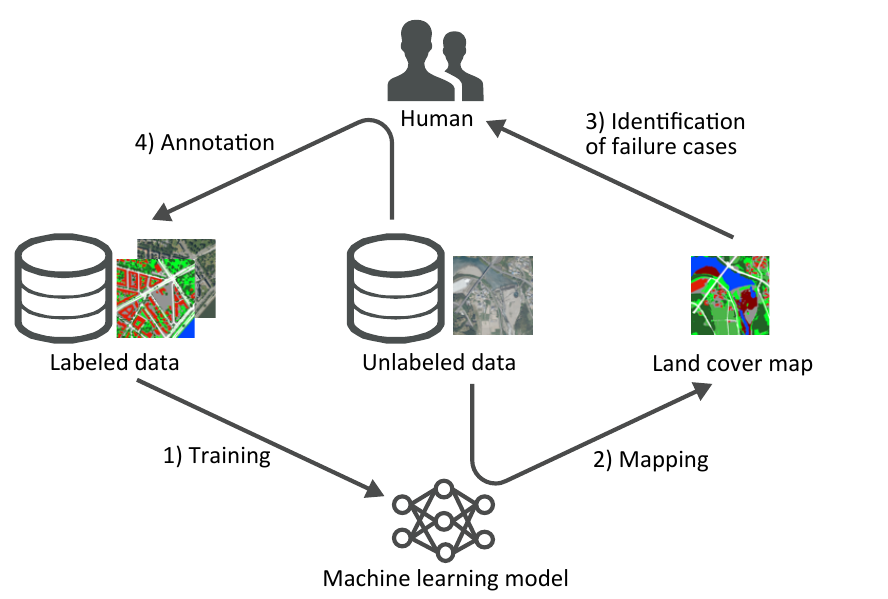}
\caption{Overview of the proposed human-in-the-loop mapping framework.}
\label{fig3}%
\end{figure}

In this paper, we propose a human-in-the-loop map creation framework that enables high-resolution mapping at a country scale by efficiently collecting additional labels necessary for mapping out-of-distribution images based on the pretrained model. The framework is an iterative process which consists of the following four steps: 1) training a land cover mapping model using labeled data, 2) mapping unlabeled data using the current model, 3) humans identifying failure cases, and 4) annotating selected unlabeled images to extend the labeled data. The overview of the framework is illustrated in Figure~\ref{fig3}. We use the OpenEarthMap dataset as the labeled data and the Geospatial Information Authority of Japan (GSI) aerial images for unlabeled data. In this study, to save computational costs, the second round of Step 2 in the cycle is the last process which was performed. This section presents the materials used for submeter-level land cover mapping in Japan. We employ public seamless aerial imagery from various data sources, which provide a comprehensive view of Japan at about 20 cm spatial resolution, although older data may be of lower quality. Handling image heterogeneity due to different acquisition times and sources is a challenge. We adopt the OpenEarthMap dataset as a basis for labeled data and make additional manual annotations to build a Japan-specific model. Government maps are used to evaluate buildings and agricultural land. Also, we describe the adopted deep learning model, the method used to select the additional unlabeled images for annotation, and the evaluation method for the mapping results.

\begin{figure}[!t]
\centering
  \subfloat[Examples of the GSI aerial imagery]{\includegraphics[width=0.9\textwidth]{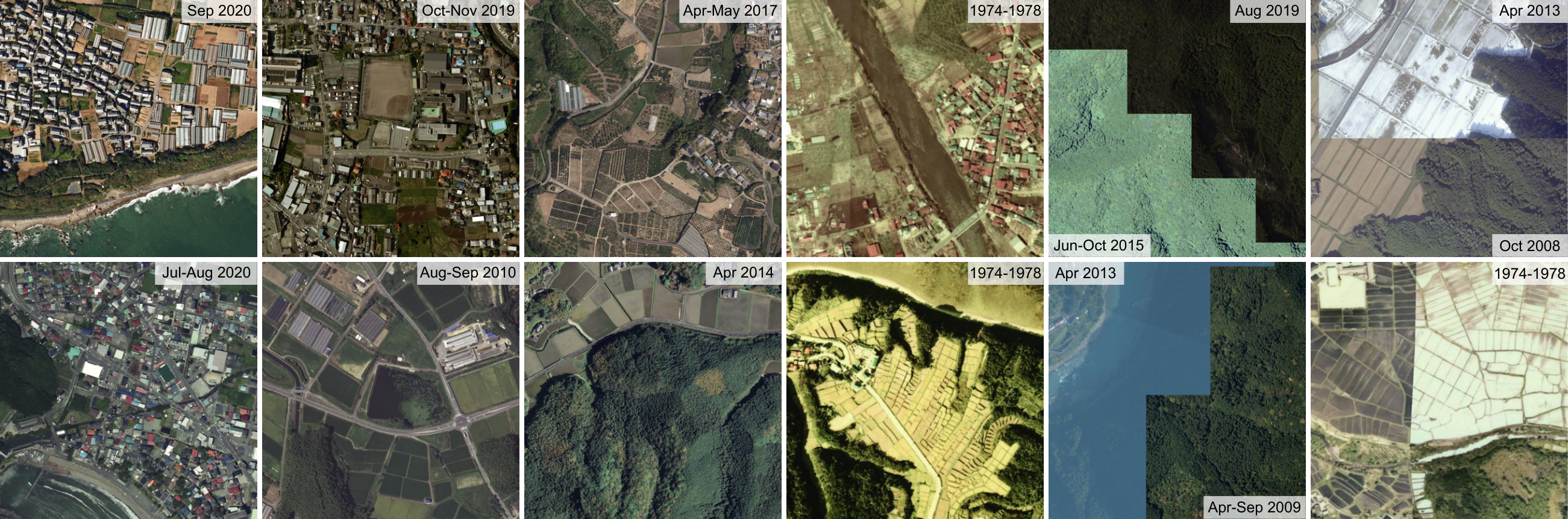}}
  \vspace{6mm}
  \subfloat[The years of aerial image acquisition.]{\includegraphics[width=0.9\textwidth]{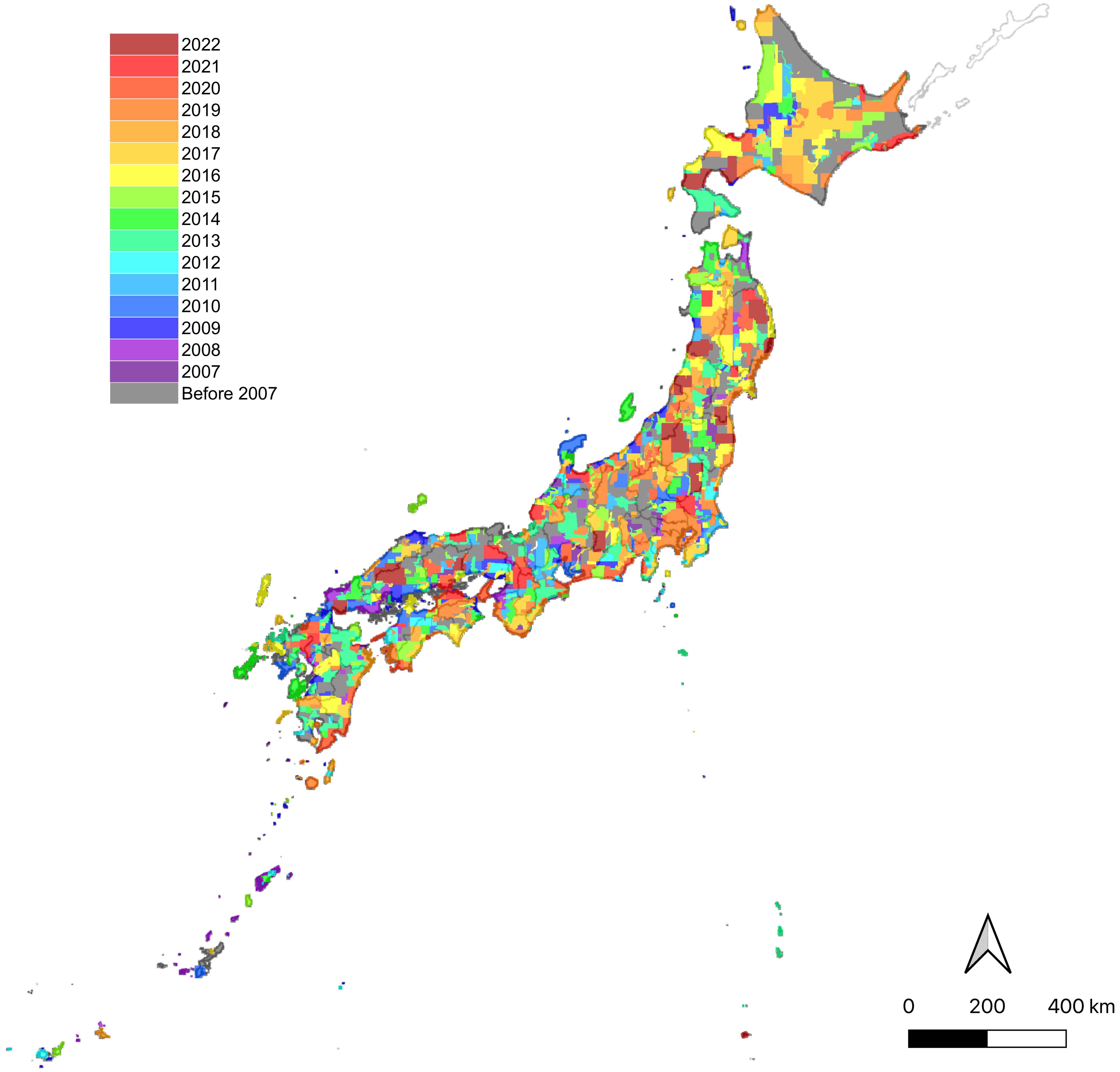}}
  \captionsetup{width=0.95\linewidth}
  \caption{The aerial photographs published by the GSI. (a) Examples of aerial imagery and (b) the imagery acquisition years.}
  \label{fig:gsi-img-yrs}
\end{figure}

\begin{figure}[!t]
	\centering 
	\includegraphics[width=\textwidth]{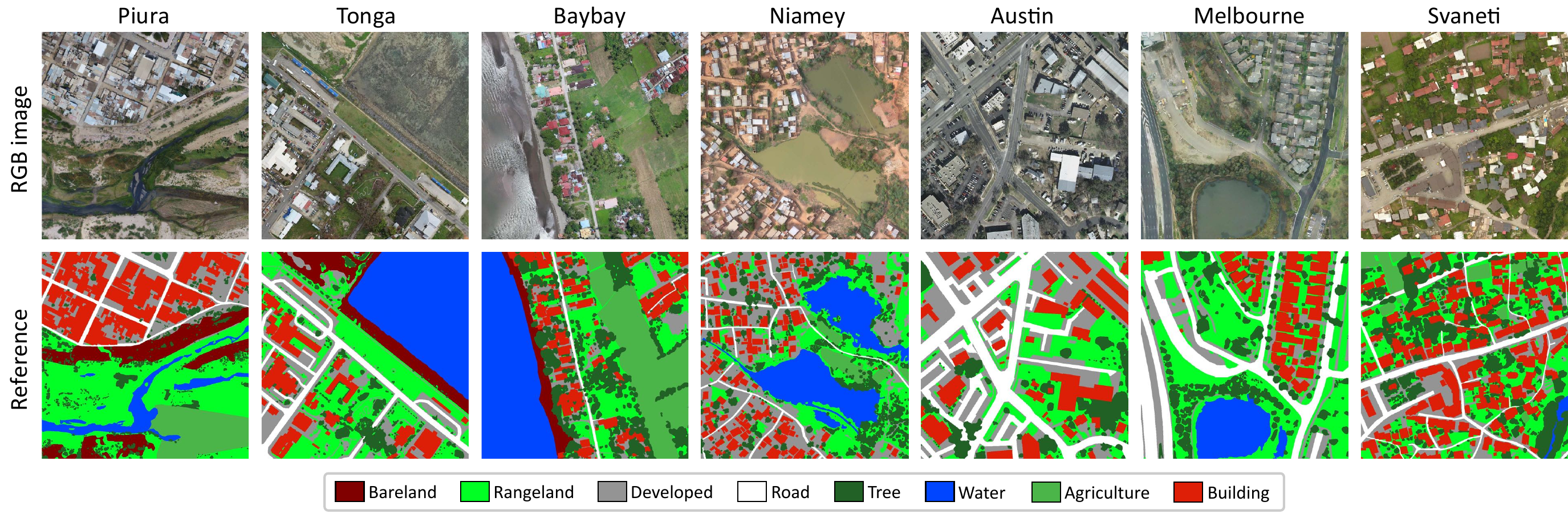}
\caption{Examples of RGB and labeled images from the OpenEarthMap data.}
\label{fig_oem}%
\vspace{5mm}
\end{figure}

\begin{figure}[!h]
	\centering 
	\includegraphics[width=0.92\textwidth]{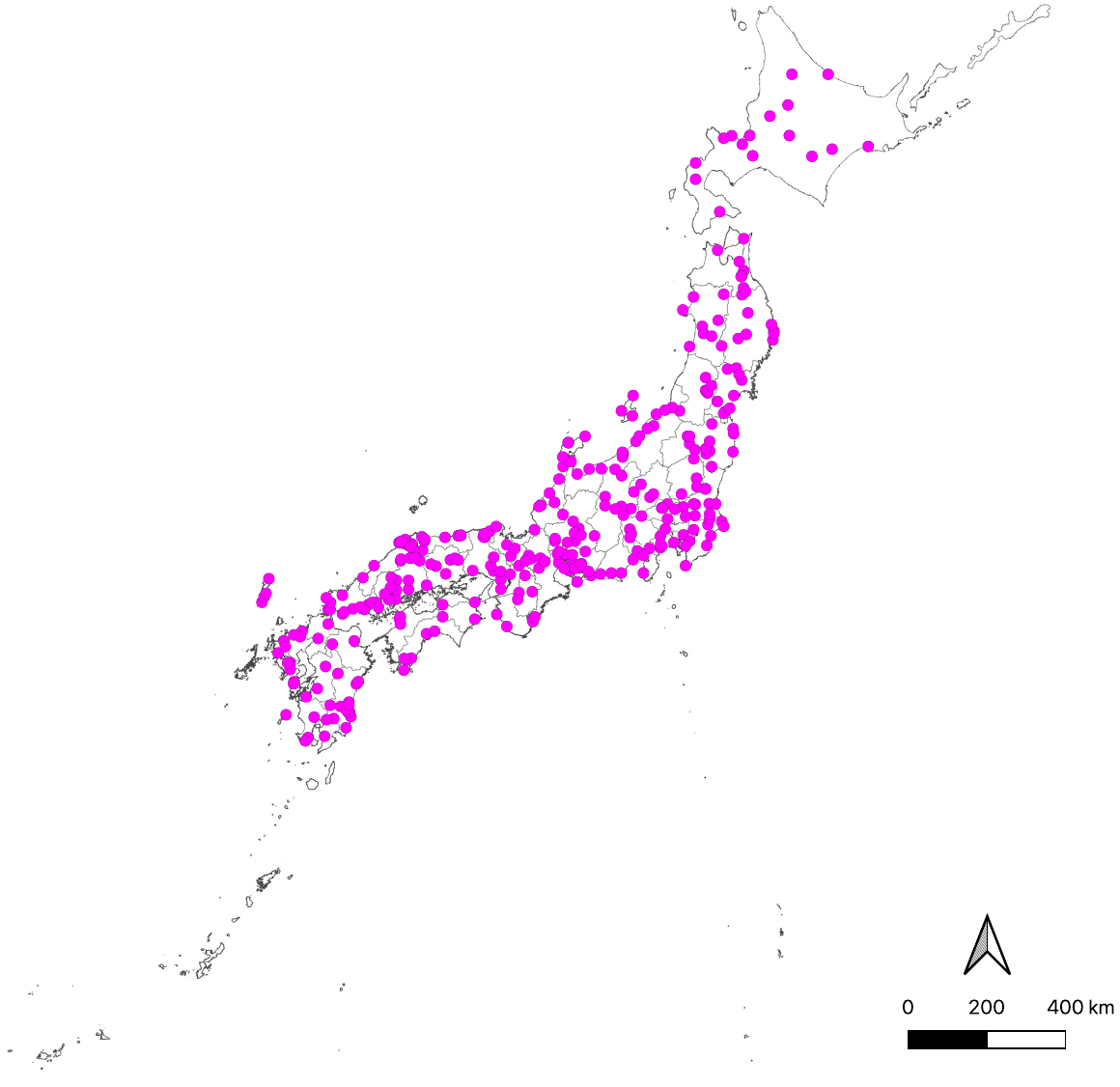}
 \vspace{-3mm}
\caption{Locations of additionally labeled aerial images.}
\label{fig2}%
\end{figure}

\subsection{Aerial Imagery}\label{sec:2.1}
Seamless aerial photographs published by the GSI\footnote{\url{https://maps.gsi.go.jp/development/ichiran.html}} were used as input images for this study. Example images are shown in Figure~\ref{fig:gsi-img-yrs}(a). The seamless aerial photographs are mosaic images of the following data sources, arranged in order of priority.
\begin{itemize}
    \item Latest orthorectified aerial images from Basic Electronic National Land Map (2007–present)
    \item Aerial photographs of national forests provided by the Forestry Agency
    \item Simplified aerial photographs (2004–present)
    \item National Land Image Information (taken from 1988 to 1990, 1984 to 1986, 1979 to 1983, 1974 to 1978)
\end{itemize}
The dates of the photographs can be checked on the web mapping\footnote{\url{https://maps.gsi.go.jp/}} provided by GSI. Figure~\ref{fig:gsi-img-yrs}(b) shows the coverage of images taken after 2007 by year in different colors, while land areas in gray are covered by images taken before 2007. The spatial resolution for all areas is about 20 cm. The image quality of very old images from National Land Image Information is very poor, as shown in the fourth column images in Figure~\ref{fig:gsi-img-yrs}(a). Because the image is a mosaic of images taken at different times (year, season, and time of day), there are many areas where the boundaries between different images are clearly visible. Such cases are shown in the fifth and sixth columns in Figure~\ref{fig:gsi-img-yrs}(a). When using OpenEarthMap for machine learning-based mapping, it is difficult to deal with these samples without fine-tuning a model because they are out-of-distribution. Ideally, more homogeneous images taken as close in time as possible should be used for land cover map product creation, but at submeter-level resolution and large scales, image heterogeneity is a real problem in many countries and regions. Under the constraints of using such heterogeneous data, this study aims to explore a framework to conduct submeter-level country-scale mapping with as few annotations as possible, using existing high-resolution land cover labels.

\subsection{OpenEarthMap Data}\label{sec:2.2}
OpenEarthMap serves as a benchmark dataset for high-resolution global land cover mapping, comprising 5,000 aerial and satellite images with manually annotated 8-class land cover labels and 2.2 million segments at a ground sampling distance of 0.25--0.5m. The dataset spans 97 regions in 44 countries, encompassing 6 continents. The 8 classes are \textit{bareland}, \textit{rangeland}, \textit{developed space}, \textit{road}, \textit{tree}, \textit{water}, \textit{agriculture land}, and \textit{building}. Each image has a size of 1024$\times$1024 pixels. Figure~\ref{fig_oem} shows seven samples of RGB and labeled images from the OpenEarthMap data. Leveraging OpenEarthMap, land cover mapping models demonstrate robust generalization capabilities worldwide and can be readily employed as off-the-shelf solutions across a diverse range of applications. OpenEarthMap contains 140 clean images from two prefectures in Japan, Tokyo and Kyoto. The data source is the GSI aerial imagery described in the Aerial Imagery section. In this study, we used the training set of the OpenEarthMap dataset, composed of 3,000 image samples. An additional 332 images with a size of $1024\times1024$ pixels were manually annotated for this study. Labeling and quality control were performed by the same annotation team that labeled the OpenEarthMap dataset, so labeling accuracy and consistency were ensured. Of the additional images, 197 (60\%) are out-of-distribution images that include wild images, and the remaining 135 (40\%) are clean images. The out-of-distribution images were sampled from areas where an OpenEarthMap pretrained model did not perform well. Figure~\ref{fig2} shows the sampling points for the 332 images. To ensure a balanced accuracy assessment, we evenly selected the additional images across different prefectures. More details of the image selection process is described in the Selection of Failures and Annotation section.
\vspace{6mm}
\begin{figure}[!h]
	\centering 
	\includegraphics[width=\textwidth]{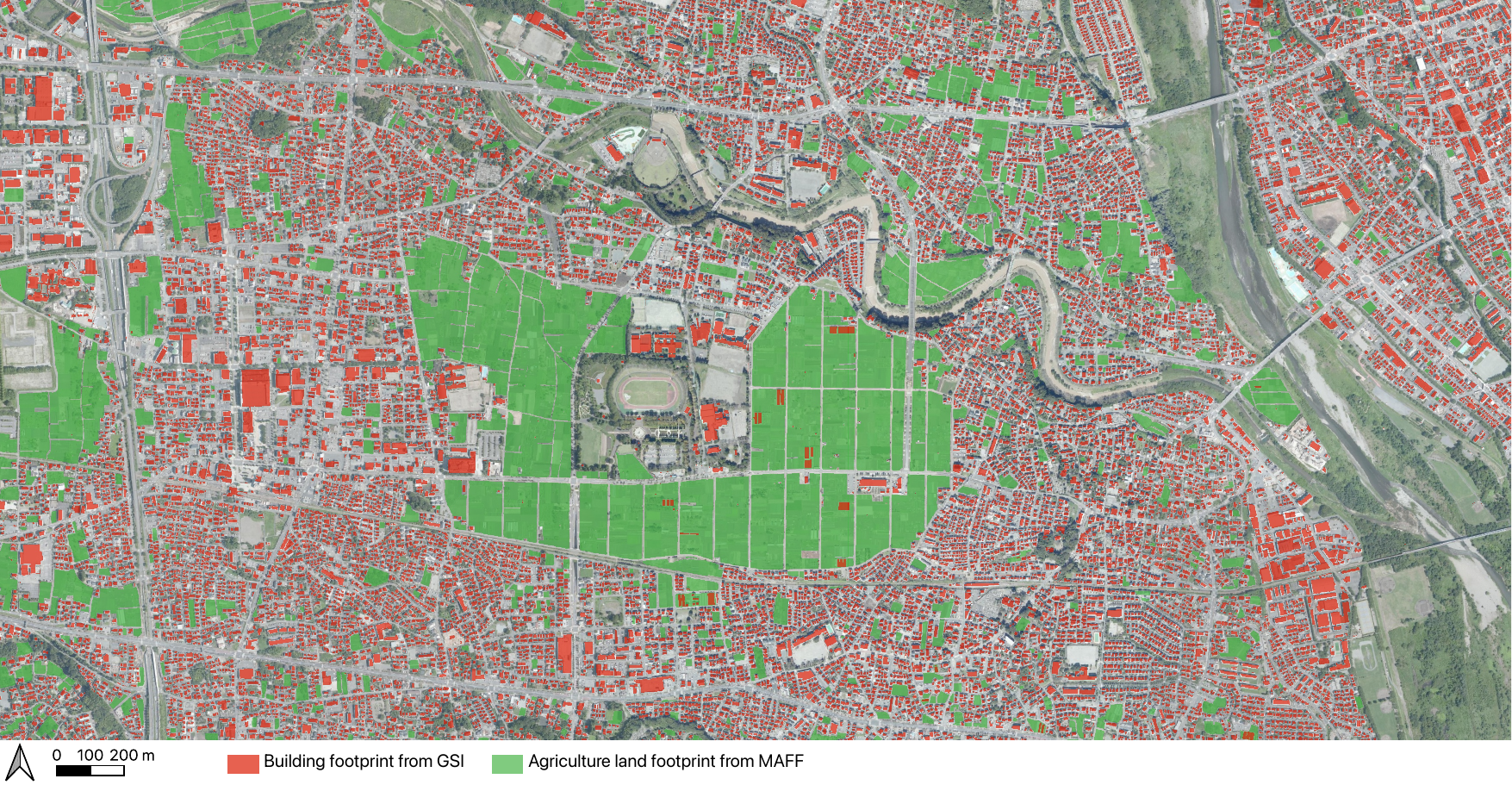}
\caption{Example of building footprints from GSI and agriculture land footprints from MAFF.}
\label{fig_ref}%
\vspace{2mm}
\end{figure}

\begin{figure}[!h]
	\centering 
	\includegraphics[width=\textwidth]{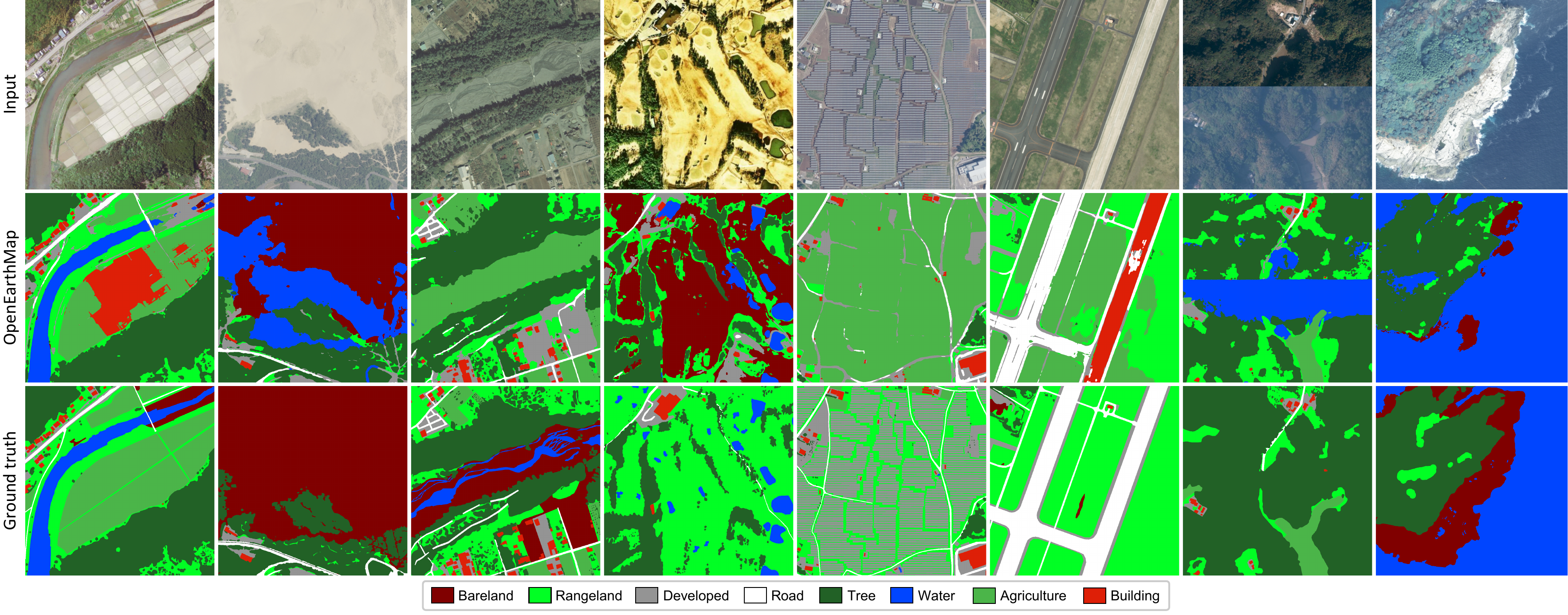}
 \captionsetup{width=0.95\linewidth}
\caption{GSI aerial images (top), failure predictions of the OpenEarthMap pretrained model (middle), and ground truth (bottom).}
\label{fig:failures}%
\end{figure}

\subsection{Reference Data}\label{sec:2.3}
For \textit{building} and \textit{agricultural land}, we use maps published by government agencies for evaluation purposes.
For building footprints, we use data distributed by GSI as part of its basic map information\footnote{\url{https://fgd.gsi.go.jp/download/mapGis.php}}. GSI provides basic map information based on survey results (e.g., city planning maps) obtained by local governments. The degree to which GSI reflects the latest information in publicly available urban planning maps depends on the local government agencies. Urban planning maps are created and managed by each local government, and the updated frequency and level of detail in the information made public can differ among municipalities and regions. For example, in Tokyo, the maps are generally updated every five years. Agriculture land footprints are obtained from the Ministry of Agriculture, Forestry and Fisheries (MAFF)\footnote{\url{https://open.fude.maff.go.jp/}}, and were manually annotated from very high-spatial resolution (0.5m) remote sensing images. Each agriculture field has a minimum size of 200 $m^2$ (400 $m^2$ for Hokkaido). The first version was completed in 2019, with subsequent annual updates. We used the version downloaded in 2022. Figure~\ref{fig_ref} shows an example of building footprints from GSI and agriculture land footprints from MAFF selected from Tokyo. These reference maps are found to be highly accurate and densely comprehensive. Note that plastic houses for agriculture are included in both reference maps, while they are labeled as \textit{agriculture land} in the OpenEarthMap dataset.

\subsection{Selection of Failures and Annotation}\label{sec:2.4}
After training the U-Net-EfficientNet-B4 model using the OpenEarthMap data as illustrated in the second step of Figure~\ref{fig3}, GSI aerial images from all over Japan were mapped with the OpenEarthMap model. In the subsequent step, the images to be annotated were selected by visually checking obvious failures from the mapping results. More specifically, the land cover mapping results were converted to the XYZ format, the mapping results were displayed on the aerial image with 70\% transparency, and the overlay was compared on and off by humans to visually search for major errors in terms of pixel count. Approximately 500 sites were identified as erroneous, of which 197 sites were selected for labeling based on geographic distribution and diversity of land cover classes. Of those 197 sites, 115 sites were labeled manually, while the labels for the remaining 82 images were corrected by a simple rule-based process using the initial mapping results (e.g., converting one class to another). In addition, within a 1$\times$1 latitude/longitude tile, we selected two images with high entropy of class probabilities based on the initial mapping results, for a total of 135 sites, and then manually annotated all of them. Finally, these 332 (i.e., 197+135) images were labeled and then divided into training, validation, and test sets, comprising 199, 33, and 100 images, respectively. Figure~\ref{fig:failures} shows eight examples of mapping failures by the OpenEarthMap pretrained model, along with manually labeled ground truth. 

\subsection{Model}\label{sec:2.5}
We used a U-Net model~\cite{ronneberger2015u} with EfficientNet-B4~\cite{tan2019efficientnet} as the backbone for mapping. This model has demonstrated both lightweight and high accuracy in comparative experiments with the latest models using the OpenEarthMap data. U-Net is a convolutional neural network architecture commonly used for semantic segmentation tasks in computer vision. It consists of an encoder and decoder with skip connections between corresponding layers, allowing the network to effectively capture both local and global features. The skip connections help preserve spatial information during downsampling and improve the network's ability to segment objects in an image accurately. Simply changing the U-Net's encoder to a more advanced backbone can lead to improved accuracy. EfficientNet is a family of convolutional neural network architectures designed to achieve better performance and efficiency by systematically scaling the model's depth, width, and resolution. It strikes a balance between accuracy and computational cost, making it highly efficient for various computer vision tasks. The EfficientNet family consists of several variants denoted by different scaling factors (B0, B1, B2, B3, B4, B5, B6, B7). We selected B4 to balance performance and computational cost. We used segmentation models PyTorch for implementation\footnote{\url{https://github.com/qubvel/segmentation_models.pytorch}}. The training was performed for 200 epochs using cross-entropy as the loss function and Adam as the optimizer. 

\subsection{Evaluation Metrics}\label{sec:2.6}
To evaluate the accuracy of land cover classification, the producer's accuracy (PA) of each class, its average accuracy (AA), and overall accuracy (OA) are used. For a given class $c \in \{1, 2, ..., K\}$, where $K$ is the number of classes ($K=8$ in this paper), we denote the number of true positives and false negatives as $\mathrm{TP}_c$ and $\mathrm{FN}_c$, respectively. PA of each class ($\mathrm{PA}_c$), AA, and OA are defined by the following formulas.

\begin{center}
    \begin{minipage}{0.75\linewidth}
        \begin{equation}
            \mathrm{PA}_c = \frac{\mathrm{TP}_c}{\mathrm{TP}_c+\mathrm{FN}_c}
        \end{equation}
        \begin{equation}
            \mathrm{AA} = \frac{1}{K}\sum_{c=1}^K \mathrm{PA}_c
        \end{equation}
        \begin{equation}
            \mathrm{OA} = \frac{\sum_{c=1}^K \mathrm{TP}_c}{\sum_{c=1}^K \mathrm{TP}_c + \sum_{c=1}^K \mathrm{FN}_c}
        \end{equation}
\end{minipage}
\end{center}
PA, AA, and OA do not consider false positives in their evaluation. To assess classification performance while taking the false positives into account, we use an additional metric known as Intersection over Union (IoU). IoU of a given class $c \in \{1, 2, ..., K\}$ is defined as:
\begin{center}
    \begin{minipage}{0.75\linewidth}
    \begin{equation}
        \mathrm{IoU}_c = \frac{\mathrm{TP}_c}{\mathrm{TP}_c+\mathrm{FN}_c+\mathrm{FP}_c},
    \end{equation}
\end{minipage}
\end{center}
where $\mathrm{FP}_c$ is the number of false positives. The mean IoU (mIoU) is defined as:
\begin{center}
    \begin{minipage}{0.75\linewidth}
    \begin{equation}
        \mathrm{mIoU} = \frac{1}{K}\sum_{c=1}^K \mathrm{IoU}_c.
    \end{equation}
\end{minipage}
\end{center}

\section{Results}\label{sec:3}
We performed both quantitative and qualitative assessments of the mapping results. For the quantitative evaluation, we employed ground truth labels from 100 manually annotated test images. In addition, we used reference data from the Geographical Survey Institute (GSI) of the Geospatial Information Authority of Japan building footprint and the Ministry of Agriculture, Forestry and Fisheries of Japan (MAFF) agriculture land footprint from five prefectures: Miyagi, Tokyo, Aichi, Osaka, and Fukuoka, considering image acquisition times and image quality. Note that we refer to the model trained on the original OpenEarthMap dataset as ``OpenEarthMap,'' and the model retrained on the extended OpenEarthMap dataset with the training set (i.e., 199 images) of the newly annotated data from Japan for this study as ``OpenEarthMap Japan''. Table~\ref{tab:pa-aa} shows class-specific producer’s accuracy (PA), its average accuracy (AA), and overall accuracy (OA), and Table~\ref{tab:iou} shows class-specific Intersection over Union (IoU) and mean IoU (mIoU) based on the test data. For all classes, significant accuracy improvement is observed through retraining, leading to an approximately 16 percent point increase in OA, reaching 80\%. Particularly, \textit{tree}, \textit{water}, \textit{agriculture land}, and \textit{building} achieve PA values exceeding 80\%. On the other hand, \textit{bareland}, \textit{rangeland}, and \textit{developed space} remain below 70\% accuracy, primarily due to ambiguous classes that are difficult to distinguish using only RGB aerial images.

\begin{table}[!t]
\captionsetup{width=0.97\linewidth}
\caption{Comparison of producer's accuracy (PA) and overall accuracy (OA) between pretrained and fine-tuning models evaluated on the test set of the OpenEarthMap Japan dataset.}
\label{tab:pa-aa}
\begin{center}
\scalebox{0.86}{
\begin{tabular}{c c c c c c c c c c c}
\hline \hline
	\multirow{2}{*}{Model}&  \multicolumn{8}{c}{PA (\%)}  & {AA} & {OA}
	\\ \cline{2-9}
	&Bareland	&	Rangeland	&	Developed	&	Road	&	Tree	&	Water	&	Agriculture	&	Building	& (\%) & (\%)  \\	\hline
OpenEarthMap       &  33.66 & 45.04 & 67.87 & 66.30 & 64.52 & 63.83 & 87.17 & 85.53 & 64.24 & 63.98 \\
OpenEarthMap Japan &  38.74 & 68.02 & 66.24 & 75.17 & 93.49 & 83.42 & 90.20 & 89.41 & 75.59 & 80.20 \\
\hline \hline
\end{tabular}}
\end{center}
\end{table}

\begin{table}[!t]
\captionsetup{width=0.98\linewidth}
\caption{Comparison of IoU between pretrained and fine-tuning models evaluated on the test set of the OpenEarthMap Japan dataset.}
\label{tab:iou}
\begin{center}
\scalebox{0.93}{
\begin{tabular}{c c c c c c c c c c}
\hline \hline
	\multirow{2}{*}{Model}&  \multicolumn{8}{c}{IoU (\%)}  & {mIoU}
	\\ \cline{2-9}
	&Bareland	&	Rangeland	&	Developed	&	Road	&	Tree	&	Water	&	Agriculture	&	Building	& (\%)  \\	\hline
OpenEarthMap       &  25.19 & 37.73 & 45.65 & 56.91 & 59.77 & 30.71 & 43.51 & 70.22 & 46.21 \\
OpenEarthMap Japan &  30.29 & 56.35 & 50.78 & 64.82 & 85.94 & 77.03 & 64.57 & 73.91 & 62.96 \\
\hline \hline
\end{tabular}}
\end{center}
\end{table}

\begin{figure*}[!t]
	\centering 
	\includegraphics[width=\textwidth]{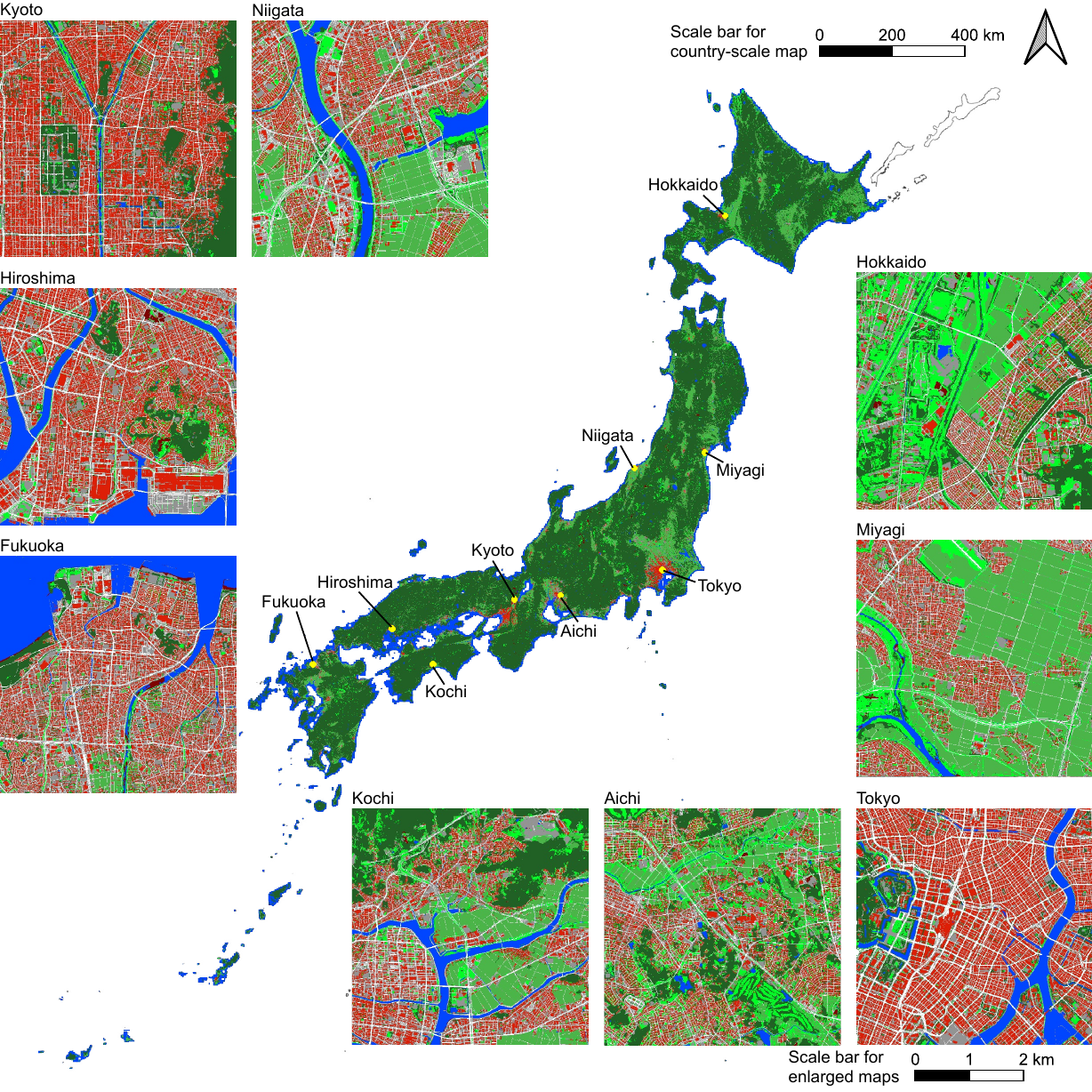}
\caption{Submeter-level land cover mapping of Japan with enlarged examples from nine prefectures.}
\label{fig:japan}%
\end{figure*}

\begin{figure}[t]
\centering
  \subfloat[OpenEarthMap model]{\includegraphics[width=0.47\textwidth]{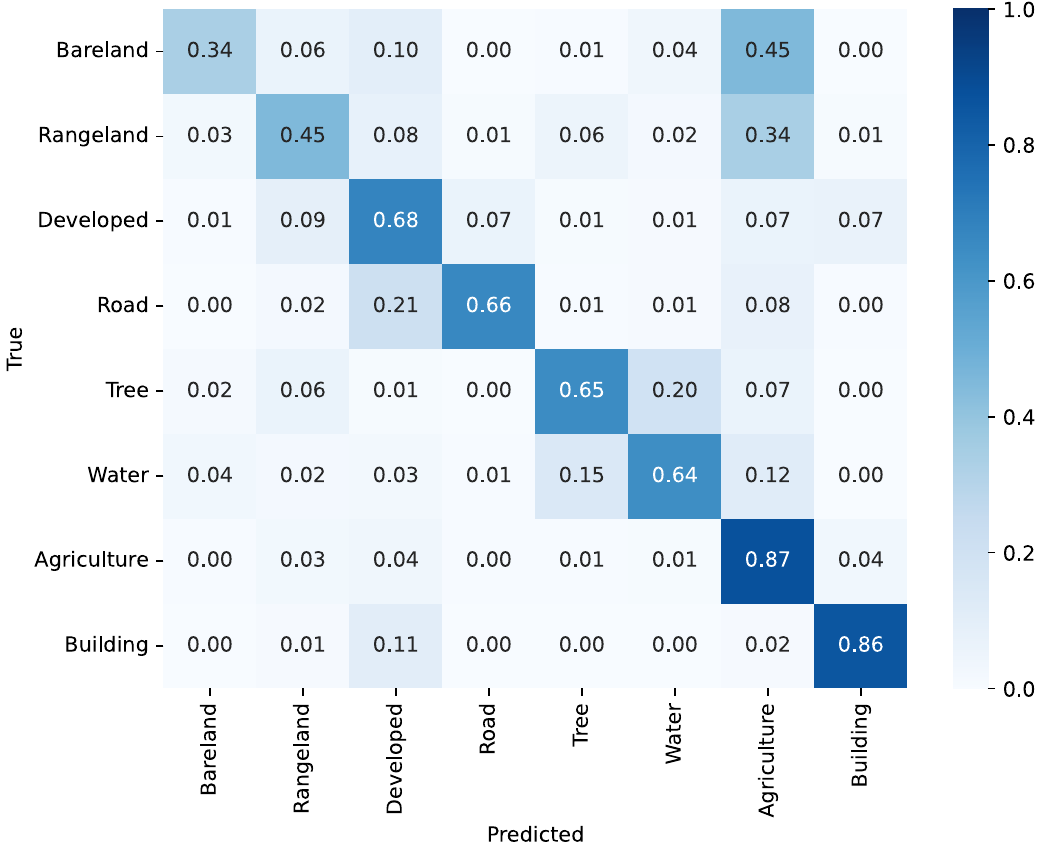}}
  \qquad
  \subfloat[OpenEarthMap Japan model]{\includegraphics[width=0.47\textwidth]{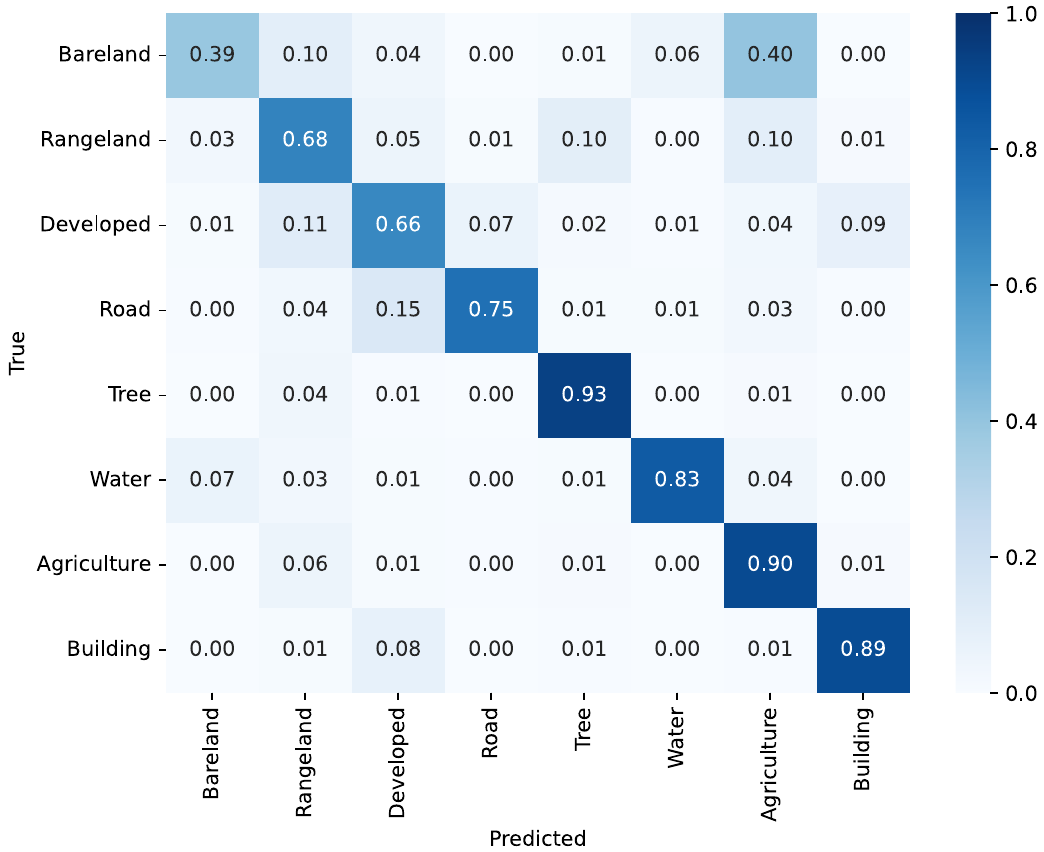}}
  \caption{Normalized confusion matrices of the (a) OpenEarthMap model and (b) OpenEarthMap Japan model.}
  \label{fig:confusionmatrix}
  \vspace{5mm}
\end{figure}

\begin{figure}[!t]
	\centering 
	\includegraphics[width=\textwidth]{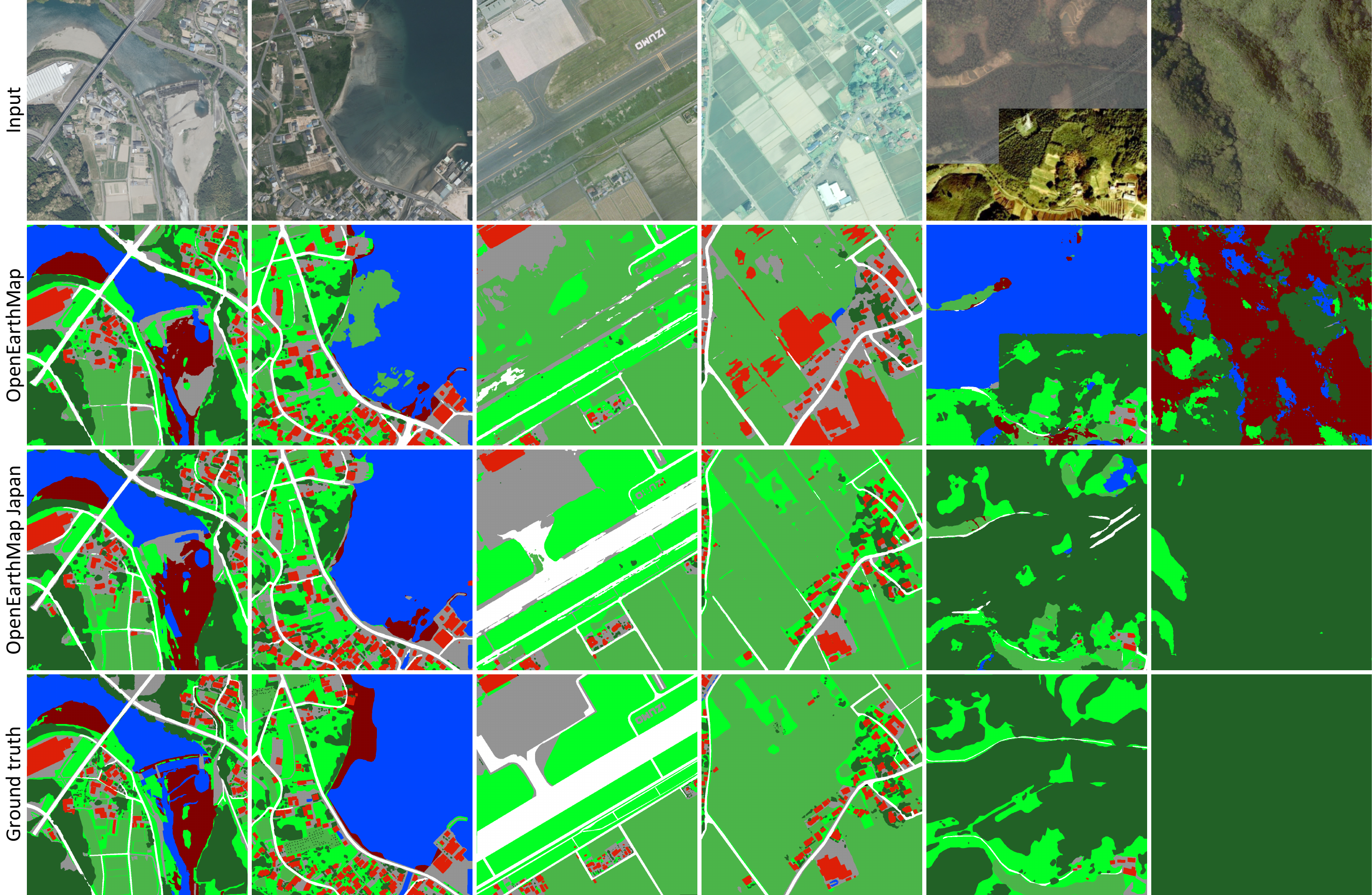}
\caption{Examples of mapping results. From top to bottom, the rows display input, results of the OpenEarthMap model, results of the OpenEarthMap Japan model, and ground truth.}
\label{fig:maps}%
\end{figure}

\begin{figure}[!t]
	\centering 
	\includegraphics[width=\textwidth]{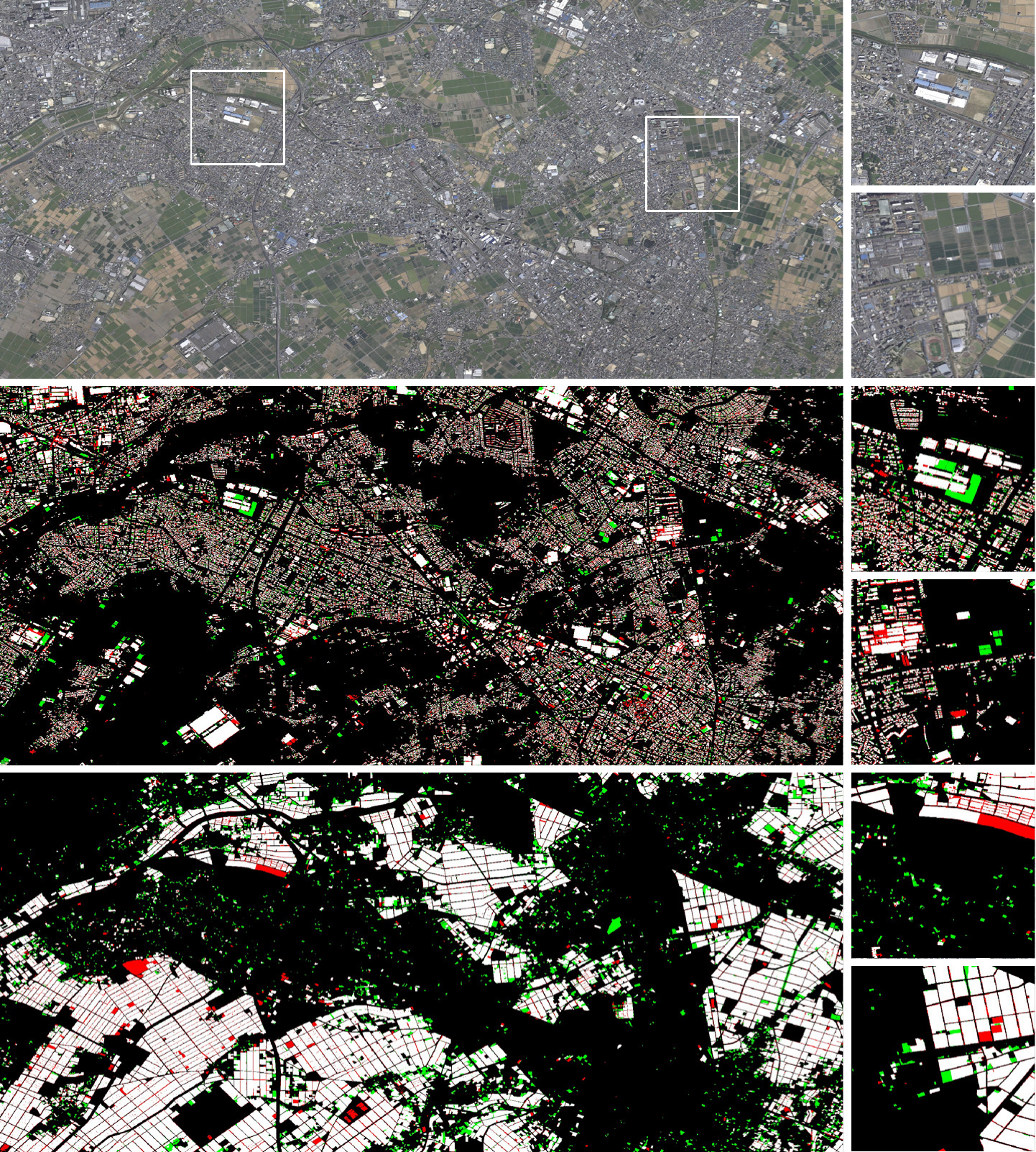}
\caption{(top) Aerial imagery, (middle) error map of buildings, and (bottom) error map of agriculture land. White, black, red, and green represent true positives, true negatives, false positives, and false negatives, respectively.}
\label{fig:error_map}%
\vspace{-4mm}
\end{figure}

Figure~\ref{fig:japan} shows the mapping result of Japan with enlarged examples from nine prefectures. It is visually evident that we were able to create highly detailed land cover maps at a submeter spatial resolution, as represented by details of roads and buildings. In the confusion matrix presented in Figure \ref{fig:confusionmatrix}, in both models over 40\% of \textit{bareland} is being misclassified as \textit{agriculture land}. In the OpenEarthMap model, approximately 34\% of \textit{rangeland} is misclassified as \textit{agriculture}, but this error is significantly reduced after retraining. Additionally, the confusion between \textit{tree} and \textit{water}, which frequently appears in forests with the OpenEarthMap model, is greatly improved after retraining. Figure~\ref{fig:maps} presents visual mapping results before and after retraining. The first column shows a clean image randomly sampled for accuracy assessment, while the remaining five columns illustrate examples where the model without retraining faced challenges. The OpenEarthMap model performs well in the first column, but in the second through sixth columns, we observe instances where it encounters challenges due to differences in data distributions. Each column highlights a typical error that arises when applying the original OpenEarthMap model to GSI aerial imagery over Japan that is out of the distribution of the original OpenEarthMap dataset. In the second column, coastal areas are incorrectly classified as \textit{agricultural land}. In the third column, we see the misclassification of \textit{rangeland} and runways (classified as \textit{road} in our scheme) at an airport as \textit{agricultural land}. The fourth column reveals the misclassification of \textit{agricultural land} as \textit{building} in brightly lit images. In the fifth column, misclassifications are evident at the boundaries where images from different time periods were stitched together. Finally, the sixth column shows significant errors in images completely covered by \textit{trees}. These errors occur because such scenes are not adequately represented in the OpenEarthMap dataset. With our addition of only 199 images for retraining, it is evident that these obvious mistakes were greatly improved.

Table~\ref{tab:reference} shows classification accuracy using the reference data distributed by the government agencies for \textit{building} and \textit{agriculture land}. For \textit{building}, we calculate metrics for the building layer of OpenStreetMap as a comparison. Retraining leads to improvements in accuracy for building, as indicated by both OA and IoU. Moreover, it demonstrates significantly better performance than OpenStreetMap, suggesting the substantial potential of large-scale mapping using deep neural networks for the automatic updating of open map data. On the other hand, regarding \textit{agriculture land}, while there is a clear improvement in IoU, the OA has decreased. This indicates a decrease in the number of true positives, but the total count of false positives and false negatives has decreased more than the decrease rate of true positives. Figure~\ref{fig:error_map} shows a GSI aerial image sampled from Aichi prefecture, containing both \textit{building} and \textit{agriculture land}, together with the error maps for these two classes. In these images, white, black, red, and green represent true positives, true negatives, false positives, and false negatives, respectively. As seen from this comparison, the main causes of false positives and false negatives in \textit{building} are 1) land use changes, 2) misalignment between reference and aerial images, and 3) prediction errors. As shown in Table~\ref{tab:reference}, the OA for \textit{building} in the manually annotated test data was nearly 90\%, indicating that approximately 10\% of the false negatives were likely due to prediction errors, while the remaining 7\% were likely attributed to changes in land cover and data misalignment. Regarding \textit{agriculture land}, as evident from the images in the bottom row of Figure~\ref{fig:error_map}, large agricultural fields are detected with high accuracy. However, smaller fields fail to be detected, as indicated by the presence of small green segments. The success or failure in detecting larger segments can be attributed to two main factors: changes in land cover and prediction errors. As shown in Table \ref{tab:reference}, in the manually annotated test data, the OA for \textit{agriculture land} reaches 90\%, suggesting that around 10\% of the false negatives can be attributed to prediction errors, while the primary factor for the remaining 11\% is likely the changes in land cover.

\begin{table}[!t]
\captionsetup{width=0.5\linewidth}
\caption{Comparison of pretrained and retrained models evaluated on the reference of building and agriculture land footprints.}
\label{tab:reference}
\begin{center}
\scalebox{1}{
\begin{tabular}{c c c c c}
\hline \hline
	\multirow{2}{*}{Model}&  \multicolumn{2}{c}{Building}  & \multicolumn{2}{c}{Agriculture}
	\\ \cline{2-5}
 & IoU & OA & IoU & OA \\ \hline
OpenStreetMap       &  35.77 &  38.69 &  --- &  ---  \\
OpenEarthMap       &  60.12 &  81.51 &  63.26 &  80.91  \\
OpenEarthMap Japan &  60.32 &  82.98 &  64.64 &  78.96  \\
\hline \hline
\end{tabular}}
\end{center}
\end{table}

\section{Discussion}\label{sec:4}
This study presents an efficient and effective approach that utilizes high-resolution OpenEarthMap data to create country-scale submeter-level land cover maps for applications such as environmental monitoring, urban planning, agriculture monitoring, and disaster management. The annotation cost (including time, labor, and computing) of remote sensing imagery poses a significant challenge in creating a country-scale land cover mapping. Adopting a human-in-the-loop framework represents a promising approach to utilize the OpenEarthMap data with a small amount of additional labeled data, to create a large-scale high-resolution land cover mapping countrywide at a low annotation cost. Here, we discuss the factors that affect the accuracy assessment, comparison with other land cover products of Japan, and the limitations and prospects of this work.

\subsection{Factors that Affect the Accuracy Assessment}\label{sec:4.1}
Accuracy assessment of land-cover products is an important step in quantifying the uncertainty of the products before they are used for related real-world applications~\cite{OLOFSSON2013122}. The accuracy assessment of the proposed OpenEarthMap Japan was based on class labels of the OpenEarthMap data, and the \textit{building} and \textit{agriculture land} footprints provided by the GSI and the MAFF, respectively, using two numerical metrics at the pixel level. Regarding the OpenEarthMap class labels, there might be some label noise that affects the metrics due to the fact that in cases where visual interpretation from single temporal imagery is hard, the disagreement between human annotators can affect the quality of the annotation (label). The lack of a classification category system that aligns the class labels of the OpenEarthMap data with that of the GSI aerial imagery data would have affected the accuracy assessment as well. Furthermore, there are time differences between the GSI images and that of the \textit{building} and \textit{agriculture land} footprints, this could have caused some land cover changes and that might affected the metrics. Another factor could be spatial misalignment between the aerial images and the footprints due to errors in the geometric correction of the aerial images.

\begin{table}[!t]
\label{tab:pa}
\captionsetup{width=0.56\linewidth}
\caption{The OpenEarthMap land cover types in relation to the JHR LULC Map version 21.03 classification categories.}
\label{tab:class-category}
\begin{center}
\scalebox{1}{
\begin{tabular}{c p{5.2cm} l}
\hline \hline
    {No.} &  {JHR LULC Map v21.03}  & {OpenEarthMap} \\
    \hline
    1  &  Water bodies & Water \\
    2  &  Built-up & Building \\
    \multirow{2}{*}{} &  & Developed space \\
       &  & Road \\
    3  &  Paddy field & \multirow{2}{*}{Agriculture land } \\
    4  &  Cropland \\
    5  &  Grassland & Rangeland \\
    6  &  Deciduous broad-leaf forest & \multirow{4}{*}{Tree} \\
    7  &  Deciduous needle-leaf forest  \\
    8  &  Evergreen broad-leaf forest  \\
    9  &  Evergreen needle-leaf forest  \\
    10  &  Bare & Bareland \\
    11  &  Bamboo forest & Tree \\
    12  &  Solar panel & Developed space \\
\hline \hline
\end{tabular}}
\end{center}
\end{table}

\subsection{Comparison with other Land Cover Products}\label{sec:4.2}
Since 2010, the JAXA Earth Observation Research Center (EORC) has been producing land use and land cover maps of Japan, called the JAXA High-Resolution Land Use and Land Cover Map (JHR LULC Map)\footnote{\url{https://earth.jaxa.jp/en/data/2562/index.html}}, at 10--50m spatial resolution using both optical sensors and synthetic aperture radar (SAR) images. The optical images are acquired by the Advanced Visible and Near Infrared Radiometer type 2 (AVNIR-2) from the Advanced Land Observation Satellite (ALOS) and some selected from Sentinel-2 L1C, while the SAR data are from ALOS-2/Phased Array Lband Synthetic Aperture Radar-2 (PALSAR-2)~\cite{takahashi2013,sota2022}. In this study, we provided a submeter-level LULC map with 8 classification categories (\textit{bareland}, \textit{rangeland}, \textit{developed space}, \textit{road}, \textit{tree}, \textit{water}, \textit{agriculture land}, and \textit{building}) at 0.25--0.5m spatial resolution using aerial images from GSI, which reflects Japan from 2007--2022 (see Figure~\ref{fig:gsi-img-yrs}), with a deep neural network model trained on the OpenEarthMap data achieving 80\% overall accuracy. In comparison, Takahashi~\textit{et al.}~\cite{takahashi2013} adopted a decision tree method to produce version 13.02 of the JHR LULC Map in 2013, reflecting Japan as of 2006--2011 at a spatial resolution of 50m with 9 classification categories (\textit{water}, \textit{urban}, \textit{paddy}, \textit{crop}, \textit{grass}, \textit{deciduous forest}, \textit{evergreen forest}, \textit{bare land,} and \textit{snow and ice}), which the overall accuracy was 89.3\%. In 2016, Sharma~\textit{et al.}~\cite{Sharma2016} produced a 30m resolution land cover map of Japan (JpLC-30) of 2013–2015 which consists of 7 land cover categories (\textit{water bodies}, \textit{deciduous forests}, \textit{evergreen forests}, \textit{croplands}, \textit{barelands}, \textit{built-up areas} and \textit{herbaceous}). The JpLC-30 map was produced with a random forests classifier which achieved an overall accuracy of 88.62\%. More recently, version 21.03 of the JHR LULC Map (10m spatial resolution) which reflects the average cover in 2018-2020 was produced by Hirayama~\textit{et al.}~\cite{sota2022} using a deep neural network method, and overall accuracy of 88.85\% was achieved in 12 classification categories (\textit{water bodies}, \textit{built-up},  \textit{paddy field}, \textit{cropland}, \textit{grassland}, \textit{deciduous broad-leaf forest},  \textit{deciduous needle-leaf forest}, \textit{evergreen broad-leaf forest}, \textit{evergreen needle-leaf forest}, \textit{bare}, \textit{bamboo forest}, and \textit{solar panel}). Because the existing LULC map products of Japan have different spatial resolutions in different classification categories, it is difficult to have a fair comparison in terms of their overall accuracy. In Table~\ref{tab:class-category}, we juxtaposed the 12 classification categories of the current JHR LULC Map version 21.03 with the 8 land cover types of the OpenEarthMap to construct semantically related classification categories among the classes of the two LULC maps.

\subsection{Limitations and Prospect}\label{sec:4.3} 
Although submeter-level land cover mapping with high accuracy is great, the limitation of this work is that the accuracy evaluation is only limited to the 8 classification categories (i.e., \textit{bareland}, \textit{rangeland}, \textit{developed space}, \textit{road}, \textit{tree}, \textit{water}, \textit{agriculture land}, and \textit{building}) of the OpenEarthMap, which fails to completely correspond to the 12 subdivision of Japan's land use and land cover types, called the JAXA High-Resolution Land Use and Land Cover Map, that has been producing by the JAXA Earth Observation Research Center since 2010 as shown in Table~\ref{tab:class-category}. We believe that the proposed framework has great potential to augment the JAXA Earth Observation Research Center map product of Japan. In future work, it is worth understanding how to verify and match the differentiated classification categories of different land use and land cover datasets, and how to extend the land types of the OpenEarthMap to all subdivisions of Japan. Also, another possibility is offered by applying the proposed framework to other countries, in particular, countries with limited available high-resolution remote sensing data such as African countries.

\section{Conclusion}
In conclusion, this work presents the first submeter-level land cover mapping of the entire country of Japan. We used aerial images provided by GSI and classified them using the U-Net-EfficientNet-B4 model, achieving an overall accuracy (OA) of 80.20\% and an average accuracy (AA) of 75.59\% across eight land cover classes. Leveraging the OpenEarthMap dataset, a benchmark for global high-resolution land cover classification mapping, we introduced a human-in-the-loop mapping framework. This framework efficiently handles challenging scenes that cannot be accurately classified by the OpenEarthMap pretrained model, requiring only a small additional label dataset. We anticipate that our framework will serve as a valuable guideline for national-scale land cover mapping.

\section*{Acknowledgments}
\par This work was supported in part by the Council for Science, Technology and Innovation (CSTI), the Cross-ministerial Strategic Innovation Promotion Program (SIP), Development of a Resilient Smart Network System against Natural Disasters (Funding agency: NIED), and JST, FOREST under Grant Number JPMJFR206S.

\section*{Data availability}
The OpenEarthMap dataset is publicly available at \hyperlink{https://zenodo.org/records/7223446}{https://zenodo.org/records/7223446}. The label data of the OpenEarthMap are provided under the same license as the original RGB images, which varies with each source dataset. For more details, please see the attribution of source data \hyperlink{https://open-earth-map.org/attribution.html}{here}. The label data for regions where the original RGB images are in the public domain or where the license is not explicitly stated are licensed under a Creative Commons Attribution-NonCommercial-ShareAlike 4.0 international license. The Geographical Survey Institute (GSI) of the Geospatial Information Authority of Japan web portal for the aerial imagery used in the study can be viewed at \hyperlink{https://maps.gsi.go.jp/development/ichiran.html}{https://maps.gsi.go.jp/development/ichiran.html}. Here, the user can find a detailed help document to navigate the site and download data. For building footprint and agricultural land footprint reference data, the GSI and the Ministry of Agriculture, Forestry and Fisheries of Japan (MAFF) distribute them via the web portals \hyperlink{https://fgd.gsi.go.jp/download/mapGis.php}{https://fgd.gsi.go.jp/download/mapGis.php} and \hyperlink{https://open.fude.maff.go.jp/}{https://open.fude.maff.go.jp/}, respectively.

\section*{Code availability}
The source code used in this study is publicly available at \url{https://open-earth-map.org/code.html}.

\bibliographystyle{unsrt}  
\bibliography{reference}  

\end{document}